\pdfoutput=1
\documentclass[10pt, logo, onecolumn, copyright]{nvidiatechreport}

\usepackage{mdframed}
\usepackage[utf8]{inputenc} 
\usepackage[T1]{fontenc}    

\usepackage{amsfonts}       
\usepackage{nicefrac}       
\usepackage{microtype}      
\usepackage[dvipsnames]{xcolor}         
\usepackage{multirow}
\usepackage{multicol}
\usepackage{graphicx}
\usepackage[numbers]{natbib}
\usepackage{tabto}
\usepackage{xspace}
\usepackage{amsmath}
\usepackage{adjustbox}
\usepackage{enumitem}
\usepackage{wrapfig}
\usepackage{dblfloatfix}

\usepackage{float}

\usepackage{natbib}

\usepackage{hyperref}
\usepackage{url}
\usepackage{booktabs}
\usepackage{multirow}
\usepackage{enumitem}
\usepackage{adjustbox}
\usepackage{tabularx}
\usepackage{arydshln}
\usepackage{wrapfig}

\usepackage{multirow}
\usepackage{colortbl}
\usepackage{amsmath}
\usepackage{makecell}
\usepackage{hhline}
\usepackage{array}
\usepackage{diagbox}
\usepackage{graphicx}
\usepackage{adjustbox}

\usepackage{tikz}
\usepackage{textcomp}
\usepackage{svg}
\usetikzlibrary{arrows.meta,fit,positioning}
\usepackage{listings}

\title{NemotronLabs VoiceChat: An Open Full-duplex Speech-to-Speech Model with Tool Calling Capabilities}

\author{
  NVIDIA
}

\begin{document}

\begin{abstract}
\textbf{Abstract.}\\
We introduce NemotronLabs VoiceChat, an open full-duplex speech-to-speech model with native tool-calling capabilities. NemotronLabs VoiceChat combines a streaming speech encoder and decoder-only language model with parallel specialized output streams for agent text and structured function calls, an auxiliary RNN-T branch for incremental user transcription, and a streaming TTS decoder. This design enables the model to listen, transcribe, reason, invoke tools, and speak within a unified streaming architecture while preserving the temporal behavior required for natural conversation. On Full-Duplex-Bench 1.0, NemotronLabs VoiceChat achieves the lowest pause-handling takeover rates among evaluated open-weight systems, 100\% takeover following user interruptions, and a 4.33/5 post-interruption response-quality score. On Full-Duplex-Bench 1.5, it resumes its response after user backchannels in 93\% of cases. NemotronLabs VoiceChat obtains a 55.1 normalized average on VoiceBench and, on Full-Duplex-Bench 3.0 (FDB 3.0), achieves 82.5\% tool-selection F1, while argument accuracy and end-to-end tool execution remain areas for improvement. These results demonstrate that full-duplex interaction, speech recognition and generation, general language capabilities, and external tool use can be integrated in a single open speech-to-speech model without sacrificing real-time conversational behavior.
\end{abstract}

\maketitle

\section{Introduction}

Speech agents are arguably a particularly effective and natural interface for interacting with intelligent systems. The conventional way of building them is through a cascaded architecture, connecting automatic speech recognition (ASR), a large language model (LLM) used in a chat function, and a text-to-speech (TTS) system  translating the agent response generated by the LLM into audio. More recent speech language models adopt a unified neural system, enabling direct speech-conditioned reasoning and speech generation~\cite{zhang2023speechgpt,fang2025llamaomni}. However, low-latency speech-to-speech generation does not reproduce the dynamics of human conversation. Most speech agents remain fundamentally \emph{half-duplex}: the system relies on a voice activity (VAD) detection module to detect when the user stops speaking, and only then begins producing its response. Human dialogue, in contrast, is inherently \emph{full-duplex}. Speakers continuously listen while speaking, take turns with fine temporal precision, produce backchannels, overlap, hesitate, and interrupt one another. Modeling these dynamics has therefore emerged as a central challenge for real-time conversational AI.

A growing body of work has established full-duplex interaction as a distinct modeling problem for speech agents. Existing approaches span explicit dialogue-state control~\cite{wang2024fullduplex}, synchronous or parallel listening--speaking architectures~\cite{veluri2024syncllm,ma2025listen}, and end-to-end models that jointly represent user and assistant audio streams~\cite{defossez2024moshi,zhang2025omniflatten,yu2025salmonnomni}. Other systems extend this landscape through modular adaptation, controllable conversational behavior, and multimodal realtime interaction~\cite{wang2025freezeomni,roy2026personaplex,cui2026minicpmo}. Collectively, these works have substantially advanced the ability of speech agents to listen and speak concurrently, respond with low latency and maintain natural conversation by handling user interruptions and backchannelling. However, their contributions are primarily centered on the dynamics of realtime interaction itself. Enabling such agents to seamlessly invoke external tools while preserving these full-duplex properties remains much less explored.

Proprietary realtime platforms already expose explicit function-calling interfaces: the OpenAI Realtime API can emit structured function calls during a realtime session, while Gemini Live similarly supports function invocation and the asynchronous return of tool results~\cite{openai2026realtime,google2025liveapi}. In open research, seamless and general tool use remains comparatively underexplored for open full-duplex speech agents. DuplexSLA\footnote{At the time of writing the model has still not been made publicly available.} addresses this problem by introducing a rate-limited textual \emph{action channel} alongside user and assistant speech, through which the model autoregressively produces planning tokens and structured actions on the same temporal timeline as the conversation~\cite{zhang2026duplexsla}. Closely related is also MoshiRAG~\cite{chien2026moshirag}, which does not perform general-purpose tool calling in the conventional sense, but demonstrates an important adjacent capability: a full-duplex speech model can detect that an utterance requires external knowledge, asynchronously trigger retrieval, and incorporate the retrieved information into its response without suspending the conversational flow.  

In this work, we introduce NemotronLabs VoiceChat, an open full-duplex speech-to-speech model with native general tool-calling capabilities. Unlike traditional cascaded stacks, this model achieves full duplex, real-time, seamless voice interaction in one unified architecture, eliminating the need for multiple models or API handoffs, thus reducing end-to-end latency. Unlike DuplexSLA, which serializes heterogeneous action-related tokens within a shared autoregressive channel, NemotronLabs VoiceChat maintains parallel, specialized streams, preserving the low-latency behavior required for full-duplex interaction. 

Our model achieves an unprecedented trade-off between general "intelligence", conversational naturalness, user-speech transcription accuracy and tool calling capabilities, while being completely open.\footnote{Checkpoint available on \href{https://huggingface.co/nvidia/NVIDIA-NemotronLabs-VoiceChat-11B}{Huggingface}.}

\section{Model Architecture}

\begin{figure}[t]
    \centering
    \includegraphics[width=1.1\linewidth]{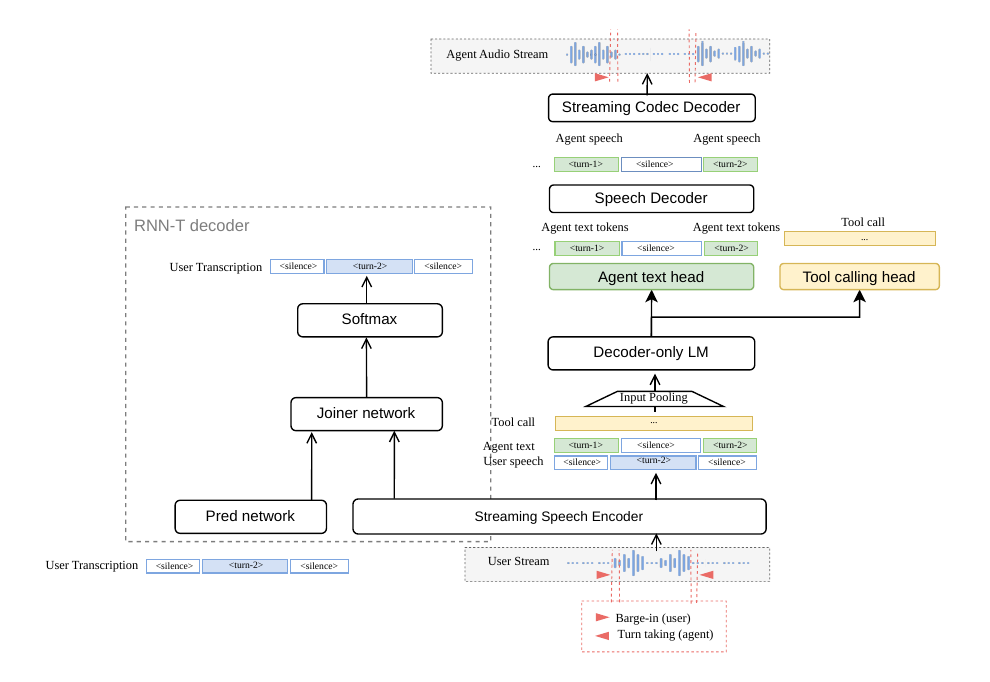}
    \caption{NemotronLabs VoiceChat Architecture Overview.}
    \label{fig:overview}
\end{figure}

Our model, depicted in Figure~\ref{fig:overview}, extends a streaming full-duplex speech-to-speech architecture with
integrated user transcription and tool-calling capabilities. A streaming speech
encoder continuously processes the user audio, while a decoder-only language
model tracks the evolving conversation and generates both the agent response
and structured function calls. The generated response is converted into speech
by a streaming speech and codec decoder, and an auxiliary RNN-T decoder predicts
the user transcription from the shared input speech representation. Together, these
components allow the model to listen, transcribe, reason, invoke tools, and
speak within a unified streaming architecture. The following subsections
describe each component in detail.
\subsection{Speech-to-Text (STT)}

The STT component comprises a perception module and a decoder-only LLM, similar to the architecture in \cite{salm_duplex, asru_duplex}. We use NVIDIA Nemotron-Nano-9B-v2-Base~\cite{nvidia2025nemotron} as the LLM backbone. The
perception module encodes streaming user speech, and the LLM predicts 
agent-text and function-call outputs. An auxiliary RNN-T branch
attached to the perception module produces incremental user transcription.
Within the perception module, an audio preprocessor converts the 16-kHz user
waveform into 128-bin log-Mel features using a 25-ms Hann window and a 10-ms
stride. These features are consumed by a 600M-parameter streaming encoder with
24 cache-aware FastConformer layers ~\cite{nvidia2026nemotron_asr_streaming} and a hidden dimension of 1,024. Causal
depthwise-striding subsampling reduces the feature sequence by a
factor of eight, producing one encoder state every 80~ms. Self-attention uses a
70-frame left context and no right context, and the convolutional modules are
also causal, so no future audio is required to produce the current state.


The perception module processes the waveform once and exposes two
representations. The raw FastConformer states are routed to the auxiliary
RNN-T branch, whereas an identity modality adapter and projection map the same
1024-dimensional states into the LLM hidden dimension. Sharing the encoder avoids running a second ASR
encoder and keeps transcription synchronized with the acoustic context used
for response generation.

The training data represents the agent response on the same 80-ms timeline as
the encoder output. Each example is initialized with padding tokens. 
For every annotated agent turn, the ordinary agent-text beginning-of-sequence (BOS) token is placed at response onset, followed by the response subword tokens in consecutive frames. When a subsequent user turn begins, the agent end-of-sequence (EOS) token serves as a stop target after a brief overlap; a final agent turn without a subsequent user turn has no EOS target.
Frames without an agent-text target, including any gap
between the last response token and EOS, remain padding. BOS and EOS therefore
serve as frame-level turn-taking targets: BOS teaches when the model should
begin responding, EOS teaches when it should stop, and padding teaches it to
remain silent. These are the normal agent-text BOS and EOS tokens and are
distinct from the function-channel boundaries \texttt{<SOTC>},
\texttt{<EOTC>}, and \texttt{<EOTR>} described in Section~\ref{sec:tool-calling-architecture}.

The raw encoder states additionally feed an RNN-T comprising a two-layer,
640-dimensional recurrent prediction network and a 640-dimensional joint
network. It predicts a 1,024-unit BPE vocabulary plus the transducer blank and
is decoded incrementally as audio frames arrive. The resulting
user transcript is exposed as an auxiliary output rather than fed into the
LLM, preserving a direct speech-conditioned response path. 

\subsection{Tool Calling}

\label{sec:tool-calling-architecture}

\begin{figure*}[t]
    \centering
    \includegraphics[width=0.98\textwidth]{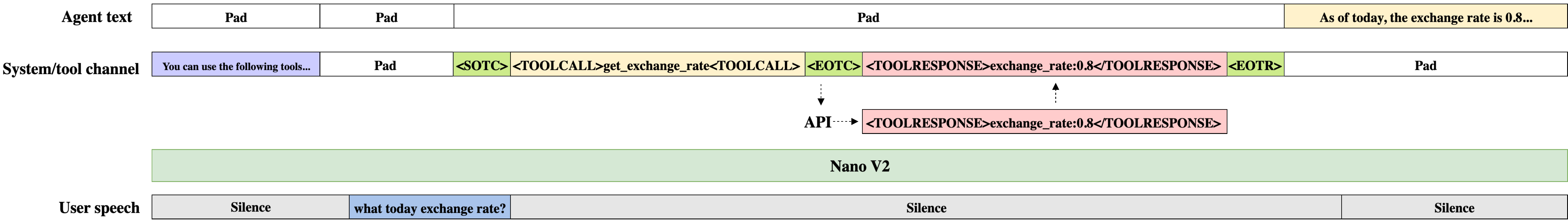}
    \caption{Tool-calling channel. For space efficiency, we omit the tool and agent-text
  channels on the input side of the figure.}
    \label{fig:tool-calling-protocol}
\end{figure*}

Tool calling is modeled with a dedicated autoregressive function channel in
parallel with the agent-text output shown in Figure~\ref{fig:tool-calling-protocol}. At
each frame, the modality-fusion layer forms a
weighted sum of the encoded user audio, the preceding agent-text-token
embedding, and the preceding function-token embedding. In our configuration,
their respective fusion weights are $1$, $1$, and $2$. The decoder-only LM
processes this fused streaming context, and a separate function head predicts
the next function-channel token. The channel emits padding while no tool action
is required; this negative supervision is important for preventing spurious
calls. When a tool is needed, the channel follows the state-machine protocol in
Figure~\ref{fig:tool-calling-protocol}.

\paragraph{Function-channel data organization.}
The data loader reads function supervision as alternating assistant-call and
tool-response segments and normalizes string-encoded arguments into JSON
objects. Each call and its corresponding response are then inserted at their
annotated frame positions, expanding the shared timeline. As illustrated in
Figure~\ref{fig:tool-calling-protocol}, \texttt{<SOTC>} (Start of Tool Call)
begins the complete tokenized call
span and \texttt{<EOTC>} (End of Tool Call) closes it; the tool-response span
follows and is terminated by \texttt{<EOTR>} (End of Tool Response). A
\texttt{<TOOLCALL>} payload is a JSON list whose elements contain a tool
\texttt{name} and \texttt{arguments}, so the same format supports one or
multiple parallel calls. Internally, \texttt{<SOTC>}, \texttt{<EOTC>}, and
\texttt{<EOTR>} denote the reserved vocabulary tokens
\texttt{<SPECIAL\_20>}, \texttt{<SPECIAL\_21>}, and
\texttt{<SPECIAL\_22>}, respectively; the textual \texttt{<TOOLCALL>} and
\texttt{<TOOL\_RESPONSE>} tags remain part of the tokenized payload. The model
is supervised on function-channel padding, the three boundary tokens, and the
call contents. Tool-response tokens are supplied as context but excluded from
the loss. During training, the corresponding positions on the agent-text
channel are supervised as padding, while user audio is replaced by silence to
keep all channels aligned. During inference, the runtime overrides this padded
agent-text interval with a predefined acknowledgement message while the tool is
executing. After \texttt{<EOTR>}, the model can resume agent-text generation or
begin another tool call, enabling multi-step tool use. The tool schema and
surface protocol are provided through the Jinja template in
Appendix~\ref{app:tool-template}.

The full training objective and token-specific loss weights are described in
Section~\ref{sec:training-objective}.

\subsection{TTS Decoder}
\label{sec:tts_decoder}

The speech-generation component of NemotronLabs VoiceChat is VoiceChat-TTS~\cite{casanova2026voicechattts}, a continuous, streamable text-to-speech decoder designed for full-duplex interaction. Unlike conventional TTS models that are invoked independently for each response, VoiceChat-TTS remains active over the complete conversation timeline and consumes the incremental assistant text-token stream produced by the upstream model. The input stream contains standard subword tokens together with three control symbols: a Beginning-of-Sequence (BOS) token marks the start of an assistant turn, padding tokens occupy intervals in which no assistant text is available, and an interruption token signals that the current utterance should stop and transition to silence. The decoder does not independently infer turn-taking from user audio; instead, it follows the text and control timeline supplied by the upstream full-duplex STT model.

VoiceChat-TTS builds on the streaming speech decoder introduced in Audio Flamingo 3-Chat~\cite{goel2025audio}. Its acoustic backbone is a 778M-parameter Gemma 3-based decoder~\cite{gemmateam2025gemma3technicalreport}, coupled with a 199M-parameter causal audio codec, for a total of 977M parameters. The codec represents 22-kHz waveforms at 12.5~Hz using 31 residual vector-quantization (RVQ) levels. Each acoustic-token frame therefore corresponds to 80~ms of waveform audio. Both the codec encoder and decoder are fully causal and cache only the convolutional history required by their receptive fields, enabling persistent streaming inference.

The acoustic backbone predicts one RVQ frame at each decoding step. Generating all 31 RVQ levels autoregressively would require 31 sequential predictions per frame; instead, we use a Mixture-of-Gaussians estimation head~\cite{kim2025efficient,goel2025audio}. The head predicts continuous representations for the remaining masked RVQ levels and progressively quantizes them over a small number of refinement iterations. In practice, 4--8 refinement iterations provide high-quality reconstruction while substantially reducing the sequential decoding depth.

Text is represented using the NVIDIA Nemotron Nano 2 subword tokenizer~\cite{nvidia2025nvidianemotronnano2}, the text stream is right-padded to the acoustic timeline, allowing the decoder to consume text incrementally while remaining active during silence intervals. We additionally offset the audio stream by one aligned decoding step relative to the text stream, providing limited linguistic look-ahead before the corresponding acoustic generation.

Directly using LLM subword embeddings can be problematic because many tokenizer units are rare or absent in TTS training data. VoiceChat-TTS therefore uses a Character-Aware Subword Encoder. Each subword is decomposed into characters and processed by a shallow Transformer encoder; the character-level representations are average-pooled to obtain the final subword embedding. A continuation embedding additionally indicates whether the current subword belongs to an ongoing lexical unit, improving pronunciation consistency for words fragmented across multiple incoming LLM tokens.

For speaker conditioning, the decoder uses a 3-second reference audio prompt. During training, the corresponding acoustic tokens prefill the beginning of the sequence, and the loss over this prompt region is masked so that the reference is used as conditioning context rather than as a reconstruction target. During inference, the same prompt initializes the acoustic context before speech generation begins. Learnable BOS and interruption embeddings provide explicit conversational boundary signals, while a gated fusion module combines text and acoustic embeddings and prevents high-magnitude RVQ representations from destabilizing mixed-precision inference.

At an interruption-token position, the model has learned to halt the current utterance and produce silence. For deterministic behavior in deployment, inference can additionally inject a fixed silence acoustic-token frame when an interruption token is received. This frame is obtained by encoding a prolonged segment of pure silence and selecting the most frequently occurring 31-token RVQ pattern~\cite{casanova2026voicechattts}. 

\section{Training Datasets and Recipes}

NemotronLabs VoiceChat is trained in two stages: continued pretraining (CPT) and supervised fine-tuning (SFT). Both stages use the same duplex data format with time-aligned user and agent streams.

\subsection{Multi-stage Training}

\paragraph{CPT.}
\label{sec:cpt}
Continual pre-training (CPT) leverages LLM pretraining corpora~\cite{su2025nemotron} to familiarize the model with speech-based language generation. We convert continuous flat text passages into pseudo-dialogues by alternating sentences between the user and agent. Turns end after one sentence with 80\% probability, with additional sentences appended at a decaying probability; a role switch is enforced once a turn exceeds 200 words. Each turn is synthesized with distinct user and agent voices, then aligned and concatenated into synchronized two-stream audio. 
During training, the loss is computed only over the agent-side text tokens in each turn using the standard next-token prediction objective, while user-side speech and previous agent text serve as interleaved context.


\paragraph{SFT.}
\label{sec:sft}

Supervised fine-tuning resumes from the continued-pretraining checkpoint and induces the conversational behavior: instruction following, turn-taking, barge-in recovery, backchannel tolerance, and tool calling. Where CPT trains on a two-component mixture of speech-text pretraining and single-turn QA data, SFT jointly
 trains on a broader collection of datasets sampled using weighted randomized round-robin, so that each capability is traded against the others under one objective. Only the full-duplex STT backbone is optimized; the audio loss weight is zero and speech is synthesized downstream by the VoiceChat-TTS decoder (Section~\ref{sec:tts_decoder}).

More details about training data construction and blending, as well as data augmentation are given in Appendix~\ref{sec:data_appendix}.




\paragraph{Component-wise training.}

The full-duplex STT backbone and streaming TTS model are trained independently.
We first optimize the full-duplex backbone through CPT and SFT to predict the
agent-text and function channels from streaming user speech. Direct audio-codec
prediction is disabled in these stages (audio-loss weight $0.0$); agent speech
is instead synthesized by the separately trained VoiceChat-TTS decoder described
in Section~\ref{sec:tts_decoder}. Consequently, gradients are not propagated
between the full-duplex backbone and TTS model.

After training the full-duplex backbone, we attach the RNN-T prediction
(decoder) and joint networks to the shared cache-aware streaming speech
encoder. We freeze the speech encoder, LLM backbone, agent-text head, function
head, and TTS model, and optimize only the RNN-T prediction and joint networks
using the standard transducer loss for user transcription. At inference time,
the independently trained full-duplex backbone, RNN-T branch, and TTS model operate together as the system shown in Figure~\ref{fig:overview}.

\subsection{Training details}

\paragraph{Training objective.}
\label{sec:training-objective}
The full-duplex STT objective combines the agent-text and function-channel
losses with the standard next-token loss on text-only data:
\begin{equation}
\mathcal{L}_{\mathrm{STT}}=(1-\lambda_{\mathrm{T2T}})
\left(\lambda_{\mathrm{text}}\mathcal{L}_{\mathrm{text}}
+\lambda_{\mathrm{FC}}\mathcal{L}_{\mathrm{FC}}\right)
+\lambda_{\mathrm{T2T}}\mathcal{L}_{\mathrm{T2T}}.
\end{equation}
For SFT, $\lambda_{\mathrm{text}}=1.0$,
$\lambda_{\mathrm{FC}}=1.0$, and $\lambda_{\mathrm{T2T}}=0.5$, giving
$\mathcal{L}_{\mathrm{STT}}=
0.5(\mathcal{L}_{\mathrm{text}}+\mathcal{L}_{\mathrm{FC}})
+0.5\mathcal{L}_{\mathrm{T2T}}$. For CPT, $\lambda_{\mathrm{text}}=3.0$,
$\lambda_{\mathrm{FC}}=1.0$, and $\lambda_{\mathrm{T2T}}=0.0$, giving
$\mathcal{L}_{\mathrm{CPT}}=3\mathcal{L}_{\mathrm{text}}
+\mathcal{L}_{\mathrm{FC}}$. Although CPT contains no tool-calling examples,
the function channel is supervised to predict padding, discouraging spurious
tool activation. Both output channels use token-weighted cross-entropy,
\begin{equation}
\mathcal{L}_{c}=-\frac{1}{N}\sum_t m_t^{(c)}w_c(y_t)
\log p_{\theta}^{(c)}(y_t\mid h_t),\qquad
c\in\{\mathrm{text},\mathrm{FC}\}.
\end{equation}
For the agent-text channel, the SFT weights are $12.5$ for beginning-of-turn,
$7.5$ for end-of-turn, $5.0$ for text content, and $1.0$ for padding; the
corresponding CPT weights are $10.0$, $10.0$, $1.0$, and $0.5$. For the
function channel, the SFT weights are $64.0$ for \texttt{<TOOLCALL>} content,
$6.0$ each for \texttt{<SOTC>} and \texttt{<EOTC>}, $3.0$ for
\texttt{<EOTR>}, and $0.3$ for padding. The function mask
$m_t^{(\mathrm{FC})}$ is zero on injected tool-response tokens, so those tokens
provide context without contributing loss. Upweighting the sparse boundary and
content tokens teaches turn-taking and the complete tool protocol, while the
padding losses discourage emissions on inactive output channels. The RNN-T and
TTS objectives are optimized separately and are not included in
$\mathcal{L}_{\mathrm{STT}}$.


\paragraph{Optimization.}

Both CPT and SFT are performed on 64 GPUs (8 nodes with 8 GPUs each) using full data parallelism and \texttt{bf16} precision. We use AdamW with $\beta_1=0.9$, $\beta_2=0.98$, zero weight decay, and a learning rate of $5\times10^{-5}$. The learning rate follows an inverse-square-root schedule with $2{,}500$ warmup steps and a minimum value of $5\times10^{-6}$. The gradient-clipping threshold is $2.0$ during CPT and $5.0$ during SFT.


\section{Inference-time Enhancements}

\paragraph{Filler messages during tool calling.} 
We have chosen a practical solution to ensure the voice agent does not remain silent, potentially for a long time, when tools are being called. For each tool a specific \textit{filler message} can be defined that will be spoken by the agent as soon as the LLM generates the text that will trigger the tool call and response. This message needs to be defined along with the tool specification in the system prompt, following the example given in Section~\ref{app:filler_messages} of the appendix. Such messages can be defined such that their duration “masks” the delay that would be experienced by the user while the tool response is being generated. For fast executing tools with short responses, they can be completely skipped. For tools with long execution time and long responses they should be long enough. A trade-off is needed in-between. 

\paragraph{Improved turn-taking.}
When the model fails to natively handle turn-taking/barge-in scenarios, we introduce a simple endpointing mechanism as a fallback, using the RNN-T transcript output. A set of heuristics based on user speech/silence activity is combined with the current response generation state, to forcefully inject BOS/EOS tokens into the model. This helps to steer the model to start a new turn or stop in case of user barge-in.

\paragraph{Optimized Inference.}
We designed an optimized inference runtime for low-latency, real-time conversation, whose details can be found in Appendix~\ref{sec:optimized_inference}.

\section{Experiments~\& Results}
We evaluate NemotronLabs VoiceChat across complementary dimensions. Full-Duplex-Bench v1~\cite{lin2025full_v1} and v1.5~\cite{lin2025full_v15} assess real-time turn management; VoiceBench~\cite{chen2026voicebench} measures single-turn response intelligence; and Full-Duplex-Bench v3 ~\cite{lin2026fdb3} evaluates tool calling under naturalistic speech conditions. These three evaluations are discussed below, while additional evaluations of speech recognition and generation quality, as well as optimized inference, are reported in Appendix~\ref{sec:further_eval}.


\paragraph{Turn-taking.}
\label{sec:fdb-turn-taking}
Full-Duplex-Bench (FDB) evaluates whether a spoken dialogue system behaves appropriately in real-time interaction. FDB~1.0~\cite{lin2025full_v1} uses pre-recorded user audio to evaluate pause handling, model backchanneling, smooth turn-taking, and user interruption. In this work, we report results for pause handling, smooth turn-taking, and user interruption. The pause tracks measure whether a model refrains from taking the floor during within-turn pauses, using both synthetic stimuli and natural pauses from CANDOR; lower Takeover Rate (TOR) is therefore preferred. The smooth-turn-taking track instead measures whether the model takes the floor after the user completes a turn. The user-interruption track measures whether the model takes the turn following an interruption, how quickly it responds, and the quality of its post-interruption response. Higher TOR is preferred on these two tracks, lower latency indicates faster interaction, and GPT-4o scores the post-interruption response quality---coherence, relevance, and adaptability---on a 0--5 scale. We adopt the official FDB scoring protocol.

For FDB~1.5~\cite{lin2025full_v15}, we evaluate the user-backchannel condition: while the model is speaking, the user produces a brief acknowledgment such as ``uh-huh'' rather than a new request. The model's subsequent behavior is classified as \emph{Respond} if it treats the acknowledgment as something to answer, \emph{Resume} if it continues its original response, \emph{Uncertain} if it expresses confusion or requests clarification, and \emph{Unknown} if it remains silent or produces an irrelevant response. Resume is therefore the desired outcome in this condition.

\begin{table*}[t]
  \caption{Full-Duplex-Bench 1.0 turn-management results. Behavioral rates are reported in percent, latency in seconds, and GPT-4o response-quality scores on a 0--5 scale. TOR denotes Takeover Rate.}
  \label{tab:fdb-1-naturalness}
  \centering
  \small
  \setlength{\tabcolsep}{3.5pt}
  \renewcommand{\arraystretch}{1.10}
  \begin{tabular*}{\textwidth}{@{}l@{\hspace{2pt}}r@{\extracolsep{\fill}}rrrrrr@{}}
    \toprule
    & \multicolumn{2}{c}{Pause handling} & \multicolumn{2}{c}{Smooth turn-taking} & \multicolumn{3}{c}{User interruption} \\
    \cmidrule(lr){2-3}\cmidrule(lr){4-5}\cmidrule(lr){6-8}
    Model & \shortstack{Synthetic\\TOR ($\downarrow$)} & \shortstack{CANDOR\\TOR ($\downarrow$)} & \shortstack{TOR\\($\uparrow$)} & \shortstack{Latency\\(s, $\downarrow$)} & \shortstack{TOR\\($\uparrow$)} & \shortstack{Response quality\\(GPT-4o, $\uparrow$)} & \shortstack{Latency\\(s, $\downarrow$)} \\
    \midrule
    \multicolumn{8}{l}{\textit{Open-weight systems}} \\
    Moshi & 98.5 & 98.0 & 94.1 & 0.265 & 100.0 & 0.77 & 0.257 \\
    Freeze-Omni & 64.2 & 48.1 & 33.6 & 0.953 & 86.7 & 3.62 & 1.409 \\
    PersonaPlex & 35.8 & 43.1 & 90.8 & 0.170 & 95.0 & 4.29 & 0.240 \\
    MoshiRAG$^{\dagger}$ & 32.0 & 56.0 & 83.0 & 0.180 & 85.0 & 3.75 & 1.020 \\
    \cmidrule(lr){1-8}
    \textbf{VC-Model} & \textbf{15.3} & \textbf{25.5} & 81.5 & 0.448 & \textbf{100.0} & \textbf{4.33} & 0.480 \\
    \midrule
    \multicolumn{8}{l}{\textit{Closed-API systems}} \\%
    Gemini Live 2.0 & 25.5 & 31.0 & 65.5 & 1.301 & 89.1 & 3.38 & 1.183 \\
    GPT-Realtime & 1.0 & 12.0 & 100.0 & 1.470 & 97.0 & 3.85 & 1.500 \\
    \bottomrule
  \end{tabular*}

  \vspace{2pt}
  \begin{minipage}{\textwidth}
    \footnotesize
    VC-Model: NemotronLabs VoiceChat. Moshi, Freeze-Omni, Gemini Live, and GPT-Realtime scores are from the public FDB results. Gemini Live 2.0 denotes the \texttt{gemini-2.0-flash-live-001} endpoint. PersonaPlex denotes the publicly released checkpoint evaluated by its authors~\cite{roy2026personaplex}. MoshiRAG$^{\dagger}$ scores are reported by its authors~\cite{chien2026moshirag}; this separate evaluation was not part of the controlled FDB run.
  \end{minipage}
\end{table*}

Among the reported open-weight systems, NemotronLabs VoiceChat achieves the lowest FDB~1.0 pause-handling TOR on both the synthetic (15.3\%) and CANDOR (25.5\%) subsets. It also reaches a 100\% user-interruption TOR and the highest response-quality score (4.33) in the comparison. Its smooth-turn TOR is 81.5\%, with smooth-turn and interruption latencies of 448 and 480~ms, respectively. PersonaPlex, however, achieves a higher smooth-turn TOR (90.8\% versus 81.5\%) and lower latency (170 versus 448~ms). On the FDB~1.5 user-backchannel condition, NemotronLabs VoiceChat achieves the highest Resume rate among the open-weight baselines: its 93\% rate exceeds the next-best result, Freeze-Omni's 80\%, by 13 percentage points, while it responds unnecessarily in only 1\% of examples.

Relative to closed-API systems, NemotronLabs VoiceChat outperforms Gemini Live 2.0 on every reported FDB~1.0 metric, including lower smooth-turn and interruption latencies by 853 and 703~ms, respectively. GPT-Realtime achieves lower pause TOR and higher smooth-turn TOR, whereas NemotronLabs VoiceChat responds faster and obtains higher user-interruption TOR and response quality. On the FDB~1.5 user-backchannel condition, NemotronLabs VoiceChat exactly matches Gemini Live 2.0 across all four behavior categories. Compared with GPT-4o Realtime, it achieves a higher Resume rate (93\% versus 70\%) and a lower Unknown rate (4\% versus 25\%).

\begin{table*}[t]
  \caption{FDB~1.5 behavioral response distribution for the user-backchannel condition. Values are percentages. A listener backchannel does not claim the floor, so Resume is the desired behavior.}
  \label{tab:fdb-15-backchannel}
  \centering
  \small
  \setlength{\tabcolsep}{3.5pt}
  \renewcommand{\arraystretch}{1.10}
  \begin{tabular}{@{}l@{\hspace{4pt}}rrrr@{}}
    \toprule
    Model & \shortstack{Respond\\($\downarrow$)} & \shortstack{Resume\\($\uparrow$)} & \shortstack{Uncertain\\($\downarrow$)} & \shortstack{Unknown\\($\downarrow$)} \\
    \midrule
    \multicolumn{5}{l}{\textit{Open-weight systems}} \\
    Moshi                    & 2.0 & 6.0  & 0.0 & 92.0 \\
    Freeze-Omni              & 7.0 & 80.0 & 2.0 & 11.0 \\
    MoshiRAG$^{\dagger}$     & 5.0 & 61.0 & 0.0 & 34.0 \\
    \cmidrule(lr){1-5}
    \textbf{VC-Model}& 1.0 & \textbf{93.0} & 2.0 & \textbf{4.0} \\
    \midrule
    \multicolumn{5}{l}{\textit{Closed-API systems}} \\
    Gemini Live 2.0           & 1.0 & 93.0 & 2.0 & 4.0 \\
    GPT-4o Realtime          & 3.0 & 70.0 & 1.0 & 25.0 \\
    \bottomrule
  \end{tabular}

  \vspace{2pt}
  \begin{minipage}{0.98\linewidth}
    \footnotesize
    VC-Model: NemotronLabs VoiceChat. Moshi, Freeze-Omni, Gemini Live, and GPT-4o Realtime values are reported by the FDB~1.5 benchmark~\cite{lin2025full_v15}. MoshiRAG$^{\dagger}$ values are reported by its authors~\cite{chien2026moshirag}; this separate evaluation was not part of the controlled FDB run. Closed-system names refer to the evaluated historical endpoints, not necessarily their current service versions.
  \end{minipage}
\end{table*}

\paragraph{Intelligence.}
\label{sec:voicebench-intelligence}
Following the evaluation of real-time turn management and overlap handling, we assess NemotronLabs VoiceChat's single-turn response intelligence with VoiceBench~\cite{chen2026voicebench}. VoiceBench tests whether voice assistants can understand spoken instructions and produce helpful, accurate, and safe responses. It combines human-recorded and synthetic speech across nine subsets spanning elementary science reasoning (OpenBookQA), multidisciplinary knowledge (MMSU), general reasoning (BBH), factual question answering (SD-QA), open-ended response quality (CommonEval, AlpacaEval-Full, and WildVoice), instruction following (IFEval), and safety (AdvBench). OpenBookQA, MMSU, and BBH use multiple-choice questions; SD-QA uses free-form responses scored against reference answers; and the three response-quality subsets are open-ended and judged on a 1--5 scale. The remaining scores use 0--100 scales. We adopt the official VoiceBench scoring protocol.

\begin{table*}[t]
  \caption{VoiceBench intelligence results for open-weight systems with full-duplex conversational capability. CommonEval (CE), AlpacaEval-Full (AE), and WildVoice (WV) are reported on a 1--5 scale; all other task scores and the normalized average are on a 0--100 scale. Higher is better. FD denotes a full-duplex speech-to-speech system, Cascade-FD denotes a full-duplex cascaded ASR--LLM--TTS system, and Omni-FD denotes an omni-modal model supporting full-duplex interaction.}
  \label{tab:voicebench-intelligence}
  \centering
  \setlength{\tabcolsep}{2.8pt}
  \renewcommand{\arraystretch}{1.08}
  \resizebox{\textwidth}{!}{%
    \begin{tabular}{lc|rrrrrrrrrr}
      \toprule
      Model & Arch. & OBQA & MMSU & CE & SD-QA & AE & BBH & WV & IFEval & AdvBench & Avg. \\
      \midrule
      Moshi & FD & 25.9 & 24.0 & 1.6 & 15.6 & 2.0 & 47.4 & 1.3 & 10.1 & 44.2 & 29.5 \\
      PersonaPlex$^\ddagger$ & FD & 24.4 & 24.9 & 2.3 & 18.8 & 2.7 & 49.4 & 2.0 & 12.0 & 8.1 & 30.6 \\
      Freeze-Omni & FD & 31.0 & 28.1 & 3.5 & 53.5 & 4.0 & 50.7 & 3.2 & 23.4 & 97.3 & 55.2 \\
      DuplexCascade & Cascade-FD & 56.0 & 52.9 & \textbf{3.6} & 45.6 & \textbf{4.4} & \textbf{59.8} & \textbf{3.6} & 43.4 & 99.0 & 65.4 \\
      MiniCPM-o 4.5$^\dagger$ & Omni-FD & \textbf{87.7} & \textbf{66.7} & \textbf{3.6} & \textbf{68.4} & 4.2 & 55.0 & \textbf{3.6} & \textbf{80.6} & 98.9 & \textbf{76.1} \\
      \cmidrule(lr){1-12}
      \textbf{VC-Model}& FD & 61.3 & 46.1 & 3.0 & 32.6 & 3.4 & 51.7 & 2.8 & 19.3 & \textbf{100.0} & 55.1 \\
      \bottomrule
    \end{tabular}%
  }

  \vspace{2pt}
  \begin{minipage}{\textwidth}
    \scriptsize
    VC-Model: NemotronLabs VoiceChat. Moshi and Freeze-Omni values are taken from the public VoiceBench leaderboard. PersonaPlex$^\ddagger$ was evaluated from its released checkpoint by the DuplexCascade authors~\cite{roy2026personaplex,yang2026duplexcascade}; these values were not reported in the original PersonaPlex paper. DuplexCascade values are reported by its authors~\cite{yang2026duplexcascade}. MiniCPM-o 4.5$^\dagger$ values are reported by the independent Raon-Speech evaluation~\cite{kim2026raonspeech}, which uses GPT-5.4 to judge the three open-ended subsets; its judge-based scores and aggregate are therefore not directly matched to the official-leaderboard evaluation.
  \end{minipage}
\end{table*}

Among open-source full-duplex models, NemotronLabs VoiceChat obtains a normalized VoiceBench average of 55.1. It substantially outperforms Moshi and PersonaPlex, improving their aggregate scores by 25.6 and 24.5 points, respectively. Its overall result is effectively tied with Freeze-Omni (55.1 versus 55.2), although the models exhibit different capability profiles. Relative to Freeze-Omni, NemotronLabs VoiceChat achieves substantially higher accuracy on OpenBookQA (61.3 versus 31.0) and MMSU (46.1 versus 28.1), and improves AdvBench safety (100.0 versus 97.3). These gains are offset by weaker performance on SD-QA, the three open-ended response-quality subsets, and IFEval. These results highlight the strength of NemotronLabs VoiceChat's 9B backbone and broader training mixture on knowledge-intensive tasks.

In the broader comparison with full-duplex systems using different architectures, the cascaded DuplexCascade system obtains an average of 65.4, while the independent Raon-Speech evaluation reports an average of 76.1 for the omni-modal MiniCPM-o 4.5. The latter uses a different judge for the open-ended subsets, as noted in Table~\ref{tab:voicebench-intelligence}, but provides a useful system-level reference point across full-duplex design choices.

\paragraph{Tool calling.}

We evaluate spoken tool use with FDB 3.0, which uses
real human speech containing disfluencies and scenarios that require chained
API calls~\cite{lin2026fdb3}.
To the best of our knowledge, our model is the first fully open full-duplex speech
model to support tool calling.
\begin{table}[t]
  \caption{FDB 3.0 tool-calling results (\%). Baseline results are from
 ~\cite{lin2026fdb3}.}
  \label{tab:fdb-3-tool-calling}
  \centering
  \begin{tabular}{lrrr}
    \toprule
    Model & Tool Sel. F1 ($\uparrow$) & Arg. Acc. ($\uparrow$) & Pass@1 ($\uparrow$) \\
    \midrule
    Ours            & 82.5 & 42.2 & 33.0 \\
    \midrule
    Gemini Live 2.5 & 78.6 & 59.3 & 49.0 \\
    Gemini Live 3.1 & 81.7 & 58.8 & 54.0 \\
    \bottomrule
  \end{tabular}
\end{table}
On FDB 3.0, the model achieves 82.5\% tool-selection F1, outperforming Gemini
Live 2.5 (78.6\%) and Gemini Live 3.1 (81.7\%) in
Table~\ref{tab:fdb-3-tool-calling}. However, its argument accuracy (42.2\%) and
Pass@1 (33.0\%) are below both Gemini Live baselines.
Because FDB 3.0 counts a sample as Pass@1 only when the model selects exactly
the expected tools and supplies perfect arguments for every call, this gap
indicates that tool routing is substantially stronger than argument extraction
and end-to-end execution. These results suggest that future improvements should
target argument grounding and multi-step tool-call composition.


\section{Conclusion}

We introduced NemotronLabs VoiceChat, an open, unified full-duplex speech-to-speech model with native tool calling capabilities. A central design choice is the use of parallel, specialized streams for agent text, function calls, and user transcription, allowing tool interaction to be integrated without serializing heterogeneous actions into the conversational output stream. Together with a streaming TTS decoder and an RNN-T branch sharing the speech encoder, this design enables the system to continuously listen, speak, transcribe, and invoke external tools while preserving the temporal structure required for natural spoken interaction.
Our evaluations demonstrate that these capabilities can be combined without sacrificing the core properties of a full-duplex voice agent as supported by our results on VoiceBench, FDB 1.0, 1.5 and 3.0, and across the OpenASR evaluation sets, while VoiceChat-TTS maintains strong intelligibility and predicted speech quality over persistent multi-turn generation. 

\section{Limitations} 
NemotronLabs VoiceChat nevertheless has several important limitations. The model is trained with audio context windows of at most approximately two minutes, and conversational information extending beyond this window may therefore not be retained reliably. Its training also explicitly balances general knowledge against conversational naturalness, transcription, turn-taking, and tool use; consequently, its knowledge, instruction-following, reasoning, and safety capabilities may be weaker than those of the underlying NVIDIA-Nemotron-Nano-9B-v2 language-model backbone. Tool use also remains imperfect: performance can degrade when many tools are exposed, with a practical recommendation of no more than five tools per session; simultaneous multi-tool invocation is not yet reliable; calls may be skipped, incorrectly selected, or supplied with invented arguments; and the model may answer from internal knowledge when a tool should instead be invoked. Long tool responses can delay subsequent speech, and user barge-in is currently unavailable while a tool is executing. Finally, robustness is limited in strongly noisy or reverberant conditions, particularly in the presence of competing background speech. 

Future work should therefore focus on extending effective conversational memory, improving argument grounding and multi-step tool composition, enabling interruption-aware tool execution, and strengthening reasoning, instruction following, and alignment without compromising real-time conversational behavior. We hope that releasing the model, training methodology, and associated resources will facilitate further research toward open speech agents that combine the fluid interaction of human conversation with reliable access to external computation and information.

\section*{Contributors}
We thank the following people for their invaluable contributions to NVIDIA NemotronLabs VoiceChat.

\noindent Jagadeesh Balam, Travis Bartley, Edresson Casanova, Sanjay Chauhan, Chen Chen, Zhehuai Chen, Zijia Chen, Francesco Ciannella, Shalini De Mello, Slyne Deng, Mikyas Desta, Harishchandra Dubey, Slim Essid, Nourchene Ferchichi, Boris Ginsburg, Mariana Graterol Fuenmayor, Negar Habibi, Kevin Hu, Anand Joseph, Viraj Karandikar, Myungjong Kim, Viacheslav Klimkov, Seelan Lakshmi Narasimhan, Lily Lee, Jason Li, Eileen Long, Ameya Mahabaleshwarkar, Aditya Malte, Adi Margolin, Amrita Mazumdar, Sasha Meister, Valentin Mendelev, Koki Nagano, Oluwatobi Olabiyi, Seonwook (Wookie) Park, Ankita Pasad, Yifan Peng, Elena Rastorgueva, Jayda Ritchie, Jason Roche, Rajarshi Roy, Nikhil Srihari, Yuanhang Su, Yoshi Suhara, Viet Anh Trinh, Jinhan Wang, Piotr Zelasko, Hui Wang, Puhui Meng, Chaosen Zhang, Yunsheng Liu, Shawn Wang, Wenjing Li, Zhonglei He.

\vfill\pagebreak

\begin{small}
\bibliographystyle{unsrt}
\bibliography{references}
\end{small}
\clearpage
\appendix

\section{Training Data Details}
\label{sec:data_appendix}

\subsection{Data construction.}
\paragraph{CPT Data Construction.} Each plain-text passage is segmented into sentences, which are alternately assigned to the user and agent to form a pseudo-dialogue. The TTS system operates in voice-cloning mode, conditioned on two distinct speaker prompts to maintain a consistent voice for each role. User and agent turns are synthesized independently, placed on a shared timeline in dialogue order, and concatenated within their respective channels. This produces synchronized two-stream audio with separate user and agent speech, which is paired with the agent-side text targets for CPT. \paragraph{SFT Data Construction.}
Like the CPT stage, the SFT stage also draws its corpora from the text-only Nemotron backbone rather than from natively recorded conversational speech, rendering text turns into speech with TTS and assembling them onto a two-channel duplex timeline in which user and agent occupy separate channels with realistic inter-turn timing. CPT trains almost entirely on TTS-rendered pretraining data, while SFT retains a large share of such data and adds instruction-tuning, conversational, and tool-calling corpora. One bucket is deliberately kept in text form, the extract-knowledge subset of the pretraining corpus, contributing a text-to-text loss at weight $0.5$ against $0$ in CPT, which preserves text-domain competence through a stage otherwise dominated by speech.

Tool-calling data is the exception, as it cannot be obtained by rendering an existing text corpus: text tool-calling transcripts contain URLs, markdown, and code that have no spoken realization, and models trained on them directly score poorly on tool relevance. These buckets are instead produced by a multi-agent pipeline that generates scenarios, turn plans, executable tool backends, and simulated conversations, followed by voice adaptation through filtering, normalization, TTS with diverse reference voices, ASR round-trip verification, and dialogue assembly.

\subsection{Data mixture}
Table~\ref{tab:mixture} reports the raw sampling weight of every bucket in both
stages. CPT assigns $0.95$ to speech-text pretraining and $0.05$ to single-turn
QA. The SFT weights sum to $1.175$ before sampler normalization: $0.55$ for
retention, $0.28$ for conversational behavior, $0.305$ for tool calling, and $0.04$ for safety. These correspond to normalized
sampling shares of approximately $46.8\%$, $23.8\%$, $26.0\%$, and $3.4\%$, respectively.

The tool-calling mixture includes multi-turn conversations that interleave
tool use, abstention, and open-domain chat. It also covers
greetings and user interruptions, teaching the model both when to invoke a
tool and when to continue the conversation without one.


\begin{table}[h]
\centering
\small
\setlength{\tabcolsep}{4pt}
\caption{Data mixture across training stages. Entries are raw sampler weights for weighted randomized round-robin fusion; a dash denotes a bucket unused in that stage. Scale is the approximate size of the shards each bucket indexes, where conv.\ denotes conversations; the pretraining bucket draws from a different collection in each stage. SFT weights are normalized by the sampler.}
\label{tab:mixture}
\begin{tabular}{@{}lp{3.2cm}cc@{}}
\toprule
Data bucket & Scale & CPT & SFT \\
\midrule
\multicolumn{4}{@{}l}{\textit{Retention}} \\
Speech-text pretraining & $\approx$530k & 0.95 & 0.40 \\
Text-only knowledge & Extract-knowledge shards (text) & -- & 0.10 \\
Spoken MCQ & $\approx$24k\,h  & -- & 0.03 \\
Single-turn QA & $\approx$12k\,h  & 0.05 & 0.02 \\
\midrule
\multicolumn{4}{@{}l}{\textit{Conversational}} \\
Duplex chat & $\approx$72k\,h / 4.3M dialogues & -- & 0.15 \\
Natural voice conversations & $\approx$4.2k\,h / 45k conv. & -- & 0.08 \\
VoiceBench-targeted & $\approx$3.8k\,h / 37k conv. & -- & 0.04 \\
Voice-agent instruction following & Instruction-following shards & -- & 0.01 \\
\midrule
\multicolumn{4}{@{}l}{\textit{Tool calling}} \\
Tool calling (all) & $\approx$7.0k\,h / 268k conv. & -- & 0.305 \\
\midrule
\multicolumn{4}{@{}l}{\textit{Safety}} \\
Spoken safety alignment & $\approx$1.1k\,h / 48k conv. & -- & 0.04 \\
\midrule
Total & & 1.00 & 1.175 \\
\bottomrule
\end{tabular}
\end{table}

\subsection{Conversational augmentation}
CPT applies only mild additive noise, since its objective is speech-text alignment rather than dialogue behavior. Turn-based data does not by itself supervise full-duplex interaction, so SFT introduces three online transformations:

\begin{itemize}
\item \textit{Early interruption} ($p=0.1$). A randomly selected agent turn is truncated mid-utterance and continues for eight further frames ($640$~ms) before EOS, so that the following user turn overlaps agent speech. Shards with pre-rendered interruptions opt out through a per-group tag.
\item \textit{Backchannel injection} ($p=0.05$ per sample, $0.5$ per agent turn). Recorded backchannel audio is loudness-matched into the user channel during agent speech, so that acknowledgements such as ``uh-huh'' are not interpreted as interruptions.
\item \textit{Text-channel delay}. During SFT, agent text targets are shifted two frames ($160$~ms) later, providing additional user audio before the model commits to each token. The function channel is not shifted, preserving the true temporal position of each tool call.
\end{itemize}

Acoustic robustness is trained in the same pass. We increase the probability
of additive DNS5~\cite{dubey2023dns5} and DEMAND~\cite{thiemann2013demand} noise from $p=0.1$ during
CPT to $p=0.5$ during SFT, using an SNR range of $-30$ to $60$~dB. SFT
additionally applies room impulse responses ($p=0.8$), microphone impulse
responses ($p=0.6$), and codec augmentation ($p=0.1$); these three
augmentations are disabled during CPT.

\section{Optimized Inference}
\label{sec:optimized_inference}

The inference runtime is designed for low-latency, real-time conversation. The perception block consisting of FastConformer encoder, processes 16-kHz input stream in streaming mode with chunk aware caching. It emits two tensors: the projected encoded audio that feeds the LLM, and the raw encoder embeddings that feeds a side-channel RNN-T. The PyTorch perception path uses CUDA Graph capture. Nemotron backbone and TTS decoders are served using a custom vLLM~\cite{kwon2023efficient} fork. The fork supports encoded-speech tensors which can be appended through an incremental interface and multi-codebook generation for TTS. Nemotron backbone has two heads: one for the agent response text and another for the function calling text. It consumes weighted sum of three channel embeddings: encoded audio, previous text token output and previous function token output.
TTS decoder auto-regressively predicts one vector of 31 audio-codebook indices per 80-ms model step using response text token and an audio prompt carrying speaker identity. A causal PyTorch codec, also accelerated with CUDA Graphs, incrementally decodes these indices into 22.05-kHz waveform samples while reusing cached state.

Parallel RNN-T decode loop consumes the encoder embeddings and tracks blank/non-blank frame density and derives begin-of-utterance, end-of-utterance and barge-in events.

Tool calls are predicted through a dedicated function channel. When this channel emits a start-of-tool-call marker, the runtime asynchronously decodes the complete call; this operation is termed "fast-decode". 
An executor then invokes the requested tool. While execution is pending, TTS and the codec synthesize a pre-configured, tool-specific acknowledgment utterance. Once the tool returns, the runtime serializes the result and forcibly inserts its tokens into the function channel and decoder context; this state update is termed "fast-inject". 
Subsequent response generation then resumes. Incoming audio may continue through the perception and RNN-T transcription path, but it is not used to condition response 
generation during tool execution. Consequently, barge-in is unavailable during this phase.

Because these stages execute in heterogeneous runtimes, a Triton Inference Server~\cite{nvidia_triton_inference_server} in Python-backend coordinates scheduling and tensor exchange among them. A FastAPI-based~\cite{ramirez_fastapi} service exposes 
a bidirectional WebSocket interface to the client. Clients send 80-ms mono PCM chunks; the server optionally resamples them to the 16-kHz model input rate and resamples the 
22.05-kHz model output for client playback. Audio is received and emitted in chunks whose durations are integer multiples of 80 ms. The input chunk duration is configurable, allowing the inference cadence to be tuned to the latency budget of a given real-time deployment.


\section{Further Evaluation}
\label{sec:further_eval}

\subsection{Safety}

We evaluate spoken safety using the AdvBench split of
VoiceBench~\cite{chen2026voicebench}, which contains 520 harmful instructions rendered
with Google TTS. Our model obtains a $100.0\%$ refusal rate
(Table~\ref{tab:voicebench-intelligence}): every evaluated response is
recognized by the official evaluator as a refusal. This result is consistent
with the explicit spoken-safety mixture used during SFT and indicates reliable,
detectable refusal behavior for the direct harmful requests covered by
AdvBench.

\subsection{Speech generation}
\label{sec:tts_results}

We evaluate the VoiceChat-TTS speech decoder independently of the upstream
full-duplex model to measure acoustic generation quality and stability under
persistent multi-turn decoding. We follow the standalone VoiceChat-TTS
evaluation protocol~\cite{casanova2026voicechattts}. The unseen-speaker
condition uses LibriTTS \textit{test-clean}, while the seen-speaker condition
uses speakers observed during training. Intelligibility is measured using word
error rate (WER), speaker preservation using speaker encoder cosine similarity
(SECS), and predicted overall speech quality using SQuIM-MOS~\cite{kumar2023torchaudio}. 

\begin{table}[t]
  \caption{TTS quality and multi-turn stability on LibriTTS. Baselines are
  evaluated on unseen speakers. VoiceChat-TTS is shown at turns 1 and 4 for
  unseen and seen speakers. Lower WER and higher SECS/SQuIM-MOS are better.}
  \label{tab:tts-quality}
  \centering
  \small
  \setlength{\tabcolsep}{3.5pt}
  \begin{tabular}{llcccc}
    \toprule
    System & Speaker & Turn
    & WER (\%) $\downarrow$
    & SECS $\uparrow$
    & SQuIM-MOS $\uparrow$ \\
    \midrule
    Ground truth
      & Unseen & -- & 1.40 & 0.830 & 4.457 \\
    Chatterbox-TTS
      & Unseen & -- & $1.24 \pm 0.02$
      & $\mathbf{0.887 \pm 0.001}$ & $4.270 \pm 0.003$ \\
    Audio Flamingo 3-Chat
      & Unseen & -- & $4.51 \pm 0.41$
      & $0.761 \pm 0.002$ & $3.600 \pm 0.009$ \\
    Qwen3-TTS-12Hz-1.7B-Base
      & Unseen & -- & $\mathbf{1.01 \pm 0.05}$
      & $0.827 \pm 0.001$ & $\mathbf{4.450 \pm 0.006}$ \\
    \midrule
    \textbf{VoiceChat-TTS}
      & Unseen & 1 & $2.00 \pm 0.10$
      & $0.757 \pm 0.004$ & $4.380 \pm 0.004$ \\
    \textbf{VoiceChat-TTS}
      & Unseen & 4 & $2.20 \pm 0.20$
      & $0.685 \pm 0.005$ & $4.376 \pm 0.003$ \\
    \midrule
    \textbf{VoiceChat-TTS}
      & Seen & 1 & $2.40 \pm 0.10$
      & $0.785 \pm 0.001$ & $4.365 \pm 0.001$ \\
    \textbf{VoiceChat-TTS}
      & Seen & 4 & $1.80 \pm 0.10$
      & $0.778 \pm 0.001$ & $4.359 \pm 0.001$ \\
    \bottomrule
  \end{tabular}
\end{table}

Table~\ref{tab:tts-quality} shows that VoiceChat-TTS retains competitive
objective speech-generation metrics while supporting the persistent decoding
behavior required by NemotronLabs VoiceChat. On the first unseen-speaker turn, VoiceChat-TTS
achieves 2.00\% WER and 4.380 SQuIM-MOS. Relative to Audio Flamingo 3-Chat,
the streaming decoder on which it is based, VoiceChat-TTS reduces WER from
4.51\% to 2.00\% and increases SQuIM-MOS from 3.600 to 4.380, with similar
first-turn SECS (0.761 versus 0.757).

Across four consecutive turns, intelligibility and predicted overall quality
remain stable for unseen speakers: WER changes from 2.00\% to 2.20\% and
SQuIM-MOS from 4.380 to 4.376. Speaker similarity, however, decreases from
0.757 to 0.685, indicating identity drift for zero-shot voices over longer
continuous contexts. For speakers observed during training, SECS remains
comparatively stable, changing only from 0.785 to 0.778 between the first and
fourth turns. This suggests that
the observed long-context speaker drift is primarily associated with zero-shot
speaker conditioning rather than a general degradation of persistent decoding.

Chatterbox-TTS and Qwen3-TTS-12Hz-1.7B-Base achieve stronger conventional
isolated-response WER, but these results do not evaluate the interaction-specific
functionality required by NemotronLabs VoiceChat. VoiceChat-TTS remains active over the
conversation timeline, generates silence under upstream PAD control, and
responds to explicit interruption signals without resetting its cached
state~\cite{casanova2026voicechattts}. End-to-end turn-taking and interruption
behavior are evaluated separately in Table~\ref{tab:fdb-1-naturalness}.

\subsection{User Transcription (ASR) Benchmark}
\begin{table}[htbp]
\centering
\caption{ASR Benchmarking results on Hugging Face OpenASR Leaderboard showing WER (\%) for all datasets~\cite{srivastav2025open}.}
\label{tab:hf-openasr-leaderboard}
\begin{tabular}{lcc}
\toprule
& \multicolumn{2}{c}{\textbf{Chunk Size (ms)}} \\
\cmidrule(lr){2-3}
\textbf{Dataset} & \textbf{80} & \textbf{160} \\
\midrule
AMI        & 15.32 & 13.54 \\
Earnings22 & 14.82 & 14.04 \\
Gigaspeech       & 12.25 & 11.65 \\
LS Clean   & 3.58  & 3.19  \\
LS Other   & 8.15  & 7.19  \\
SPGISpeech       & 3.91  & 3.56  \\
Tedlium   & 5.25  & 4.89  \\
VoxPopuli  & 8.85  & 8.17  \\
\midrule
Average    & 9.02  & 8.28  \\
\bottomrule
\end{tabular}
\end{table}

We evaluated the proposed model on the Hugging Face OpenASR Leaderboard datasets~\cite{srivastav2025open} which is a standardized benchmark for comparing automatic speech recognition systems across diverse speech domains and acoustic conditions. Following the leaderboard protocol, we report word error rate (WER) on AMI, Earnings22, GigaSpeech, LibriSpeech clean and other, SPGISpeech, Tedlium, and VoxPopuli which were computed using the scoring scripts from the leaderboard. To study the trade-offs between streaming latency and transcription accuracy, we evaluated the proposed model with two chunk-size specifically 80~ms and 160~ms. A smaller chunk-size reduces the amount of audio that must be observed before the model produces the user transcription, thereby enabling lower-latency streaming ASR. In contrast, a larger chunk-size provides additional local acoustics and linguistics context which improves the accuracy of decoded words. As shown in Table~\ref{tab:hf-openasr-leaderboard}, increasing the chunk-size from 80~ms to 160~ms consistently improved the ASR performance across all the evaluation sets reducing the average WER from 9.02\% to 8.28\%. This suggests the benefits offered by long chunk-size in accurate transcription of noisy long-form speech in challenging conversation environments.

Importantly, \texttt{nemotron-speech-streaming-0.6b}~\cite{nvidia2026nemotron_asr_streaming} uses a cache-aware streaming architecture that seamlessly supports multiple chunk-size specified in the model training configuration. Consequently, the same trained model can be deployed under different latency requirements without the need to retrain separate models for different latency. This flexibility enables practitioners to select a smaller chunk-size when responsiveness is critical, or a large chunk-size when ASR accuracy is preferred with the same unified streaming model.

\subsection{Inference efficiency}

All inference measurements for the baseline system were conducted on a single NVIDIA H100 PCIe GPU with 80 GB of memory. The perception encoder and LLM backbone run in BF16, whereas the TTS backbone and cached Mamba recurrent states remain in FP32. On supported GPUs, TF32 execution is enabled for eligible FP32 matrix multiplications. Lower-precision and quantized variants were not evaluated. With four concurrent streams, the per-stream p95 inference latency is 118 ms per 160-ms audio chunk. This corresponds to 1.36× real-time processing throughput.

\pagebreak

\section{Tool Specification}

\subsection{Tool-Calling Filler Message Specification}
\label{app:filler_messages}
Example of tool definition with filler message:

\begin{lstlisting}[
  basicstyle=\ttfamily\scriptsize,
  breaklines=true,
  columns=fullflexible,
  keepspaces=true,
  showstringspaces=false,
  frame=single,
  label={lst:tool-calling-filler-message}
]
{
    "name": "calculate_bmi",
    "description": "Calculate the Body Mass Index (BMI)",
    "parameters": {
      "type": "object",
      "properties": {
        "weight": {
          "type": "number",
          "description": "The weight in kilograms"
        },
        "height": {
          "type": "number",
          "description": "The height in meters"
        }
      },
      "required": [
        "weight",
        "height"
      ]
    },
    "ack_message": "Sure, let me calculate that for you"
  }
\end{lstlisting}

\textbf{Quick Example} \\

\noindent
\textbf{User (spoken):} ``What's the weather in Tokyo?''

\medskip
\noindent
$\rightarrow$ \textbf{Model generates:} \\
\texttt{\detokenize{[{"name": "get_weather", "arguments": {"city": "Tokyo"}}]}}

\medskip
\noindent
$\rightarrow$ \textbf{Agent speaks \texttt{ack\_message}:}\\
``Let me check the weather for you.''

\medskip
\noindent
$\rightarrow$ \textbf{Tool returns:} \\
\texttt{\{"temperature": "22\textdegree{} C", "condition": "Sunny"\}}

\medskip
\noindent
$\rightarrow$ \textbf{Agent speaks:}
``It's 22 degrees and sunny in Tokyo right now.''

\subsection{Tool-Calling Jinja Template}
\label{app:tool-template}

The following Jinja template renders the system prompt used by the
speech-to-speech backend. When tool definitions are available, it serializes
their schemas inside \texttt{<AVAILABLE\_TOOLS>} and appends the expected
\texttt{<TOOLCALL>} and \texttt{<TOOL\_RESPONSE>} surface formats. The user
turn is supplied as audio and is therefore not rendered by this template.

\begin{lstlisting}[
  basicstyle=\ttfamily\scriptsize,
  breaklines=true,
  columns=fullflexible,
  keepspaces=true,
  showstringspaces=false,
  frame=single,
  caption={Jinja template for rendering the tool-enabled system prompt.},
  label={lst:tool-calling-jinja-template}
]
{#-
    Voicechat tool-calling chat template for Nano v2.

    Adapted from the full nano_v2_chat_template.jinja released with Nemotron Nano v2
    (see https://huggingface.co/nvidia/NVIDIA-Nemotron-Nano-9B-v2).
    This template renders only the system prompt (with tool definitions
    injected), since the s2s_voicechat backend receives audio as user
    input and tokenises the system prompt separately.

    Variables:
      - system_message (str): original system prompt text.
      - tools (list|None): OpenAI-format tool definitions.
-#}
{{- system_message -}}
{%- if tools -%}
    {%- if system_message != '' -%}
        {{- '\n\n' -}}
    {%- endif -%}
    {{- 'You can use the following tools to assist the user if required:' -}}
    {{- '\n<AVAILABLE_TOOLS>[' -}}
    {%- for tool in tools -%}
        {%- set _t = (tool.function if tool.function is defined else tool) -%}
        {%- set _d = {} -%}
        {%- for k, v in _t.items() if k != 'type' -%}
            {%- set _ = _d.update({k: v}) -%}
        {%- endfor -%}
        {{- _d | tojson -}}
        {{- ', ' if not loop.last else '' -}}
    {%- endfor -%}
    {{- ']</AVAILABLE_TOOLS>\n\n' -}}

    {{- 'If you decide to call any tool(s), use the following format:\n' -}}
    {{- '<TOOLCALL>[{"name": "tool_name1", "arguments": "tool_args1"}, ' -}}
    {{- '{"name": "tool_name2", "arguments": "tool_args2"}]' -}}
    {{- '</TOOLCALL>\n\n' -}}

    {{- 'The user will execute tool-calls and return responses from tool(s) in this format:\n' -}}
    {{- '<TOOL_RESPONSE>[{"tool_response1"}, {"tool_response2"}]</TOOL_RESPONSE>\n\n' -}}

    {{- 'Based on the tool responses, you can call additional tools if needed, correct tool calls if any errors are found, or just respond to the user.' -}}
{%- endif -%}
\end{lstlisting}

\end{document}